# Curriculum-guided multimodal representation learning enables generalizable prediction of nanomaterial-protein interactions

Hengjie Yu[1,2], Kenneth A. Dawson[3], Haiyun Yang[1], Shuya Liu[1], Yan Yan[3,4], Yaochu Jin[1,2]*

[1]School of Engineering, Westlake University, Hangzhou, Zhejiang 310030, China

[2]Institute of Advanced Technology, Westlake Institute for Advanced Study, Hangzhou, Zhejiang 310024, China

[3]Centre for BioNano Interactions, School of Chemistry, University College Dublin, Belfield, Dublin 4, Ireland

[4]School of Biomolecular and Biomedical Science, UCD Conway Institute of Biomolecular and Biomedical Research, University College Dublin, Belfield, Dublin 4, Ireland

* Corresponding author: Yaochu Jin. Email: jinyaochu@westlake.edu.cn

## Abstract

Nanomaterial-protein interactions (NPI) are pivotal to realizing the therapeutic and diagnostic potential of nanomaterials. Although AI promises to accelerate mechanistic understanding and enable rational nanomaterial design, robust generalization to unseen nanomaterials or proteins remains unresolved. Here, we present CuMMI (curriculum-guided multimodal interaction model), a generalizable, explainable, and transferable model designed to infer NPI across complex biological settings. CuMMI leverages a self-constructed million-scale NPI dataset and adopts a multi-stage curriculum centered on human plasma, with progressively broader biofluid exposure to enhance data coverage and generalizability. By integrating protein sequence, structure, and a text-encoded experimental context of 37 features, CuMMI captures complementary material-specific, biochemical, and environmental information. Sample-level quality weights are assigned to ensure full utilization of available data while mitigating low-confidence and sparsely recorded entries. Ablation studies highlight the most influential tabular features, clarifying their contribution to the prediction. Through rigorous external validation across independence-preserving temporal, nanomaterial-held-out, and protein-held-out evaluations, our framework consistently achieves good performance (mean of five classification metrics exceeding 0.75), highlighting its robustness and generalizability to unseen data. Furthermore, fine-tuning on independent gold-nanoparticle data and a held-out protein subset further delivers better performance than training from scratch with substantially fewer samples. Together, our approach enables generalizable and transferable NPI prediction and may accelerate *in vitro* research and applications of nanomaterials.

## Main

Nanomaterials have attracted significant attention due to their unique physical, chemical, and biological properties, playing an increasingly important role across diverse fields, including biomedicine[1], agriculture[2,3], environment[4,5] and beyond. Protein interactions with surfaces, including binding and unfolding, have been studied since the 1960s[6]. Understanding the interactions between nanomaterials and biological environments is essential for advancing their efficient and sustainable applications. Among these interactions, the spontaneous formation of the protein corona on nanomaterial surfaces can profoundly alter their physicochemical properties and biological behaviors[7]. However, the interaction between nanomaterials and protein is influenced by a multitude of factors[8,9], including the physicochemical properties of nanomaterials (such as size, surface charge, and concentration), the type of biological fluid involved (e.g., plasma, serum, or cerebrospinal fluid from human or mouse), and environmental parameters where nano-bio interactions happen (such as temperature, flow rate, and incubation time). In addition, protein corona isolation and proteomic characterization are highly sensitive to the experimental protocols and analytical methodologies employed[10,11]. Given these complexities, experimental studies involving labor-intensive and costly characterization are fundamentally constrained in their ability to systematically probe multiple variables. These limitations highlight the need for scalable, data-driven approaches to elucidate the principles underlying protein corona formation and to enable accurate prediction of nanomaterial-protein interactions (NPI).

Artificial intelligence (AI), with their capacity to model complex, nonlinear relationships, holds great promise for predicting protein corona composition. A summary of main AI applications (2018-2025) is given in Supplementary Table S1. Early in 2018, one of the first machine learning (ML) models applied random forest (RF) classification to predict protein corona formation on silver nanoparticles, achieving high accuracy (F1 score = 0.81) based on nanoparticle, protein, and solution properties[12]. In 2020, a study built a dataset of 652 nanoparticle-protein interactions and developed 178 RF regression models to predict the relative protein abundance (RPA) of each protein, reaching correlation coefficient above 0.75[13]. Recent research further leveraged this dataset to train traditional ML models capable of predicting RPA, and deployed the models[14] and dataset[15] online available. In addition to predicting the affinity of specific proteins for different nanomaterials, RF classifiers have also been employed to predict the binding affinity of various proteins toward a given type of carbon nanotube, achieving an accuracy of 0.78[16]. Recently, deep learning, based on the combination of graph neural network and multilayer perceptron, was used to predict NPI between gold nanoclusters and blood proteins[17]. However, these existing ML prediction tools are constrained by limited datasets, representation capacity, and model architectures, typically offering predictive capabilities for only specific nanomaterials or proteins. This inherent lack of generalization to previously unseen nanomaterials or proteins limits the broader applicability and practical utility of such models.

Recent advancements in pre-trained models, such as protein language models (e.g., ESM2[18] and AlphaFold 3[19]) and text embedding models (e.g. Linq-Embed-Mistral[20] and BERT[21]), offer unprecedented opportunities to develop generalizable predictive frameworks. Protein language models have been applied to predict structure[22], function[23], and molecular interactions[19] and to generate protein sequences[24]. In contrast to conversational models like ChatGPT, text embedding models are specifically designed to generate dense numerical representations that capture the semantic content of entire texts, enabling efficient similarity comparisons and downstream learning, such as molecule entity understanding[25] and materials discovery[26]. Trained on extensive and diverse datasets, these pre-trained models can learn unified, transferable representations that capture intricate semantic relationships, laying a robust foundation for generalizable learning. In parallel, curriculum learning offers a complementary route to generalization by modulating how a model is exposed to data: it mimics human learning by presenting examples in a structured order, typically progressing from simpler or higher-confidence instances to harder and more distribution-shifted cases[27]. Such staged training can improve robustness and transfer by gradually broadening coverage while reducing early overfitting to noise or spurious correlations[28]. We therefore hypothesize that, given sufficiently comprehensive datasets, combining pretrained multimodal representations with curriculum learning can enable generalized prediction of NPI: pretraining provides a broadly transferable representational foundation, whereas curriculum learning provides a high-level learning strategy that progressively broadens the model's experience.

The contributions of this work, aimed at enhancing the generalization of NPI prediction models, are summarized in Figure 1. We construct the largest curated NPI dataset to date (Figure 1a), including 1.97 million samples and 37,392 proteins. Biofluids are organized into four source categories, and each sample is assigned a composite quality weight to support robust learning from heterogeneous literature data. We propose CuMMI (curriculum-guided multimodal interaction model), a generalizable and transferable NPI prediction framework that leverages curriculum-based learning strategies (Figure 1b) together with the intrinsic generalization of multimodal representations (Figure 1c) to achieve robust generalization across data distribution shifts. Beyond predictive performance, we provide ablation-based explainability (Figure 1d) through modality and tabular-feature ablations, including single and pairwise feature/group analyses, to quantify contributions and interactions, helping build trust and offering actionable insight. Besides, we validate model generalization using three strictly independent external evaluations (Figure 1e), including out-of-distribution (OOD) splits by timeline, nanomaterial, and protein. Additionally, fine-tuning on independent datasets further demonstrates the potential of the model as a foundational model across different domains (Figure 1f). This work lays the foundation for generalizable NPI predictions, with significant implications for advancing biomedical research and therapeutic applications.

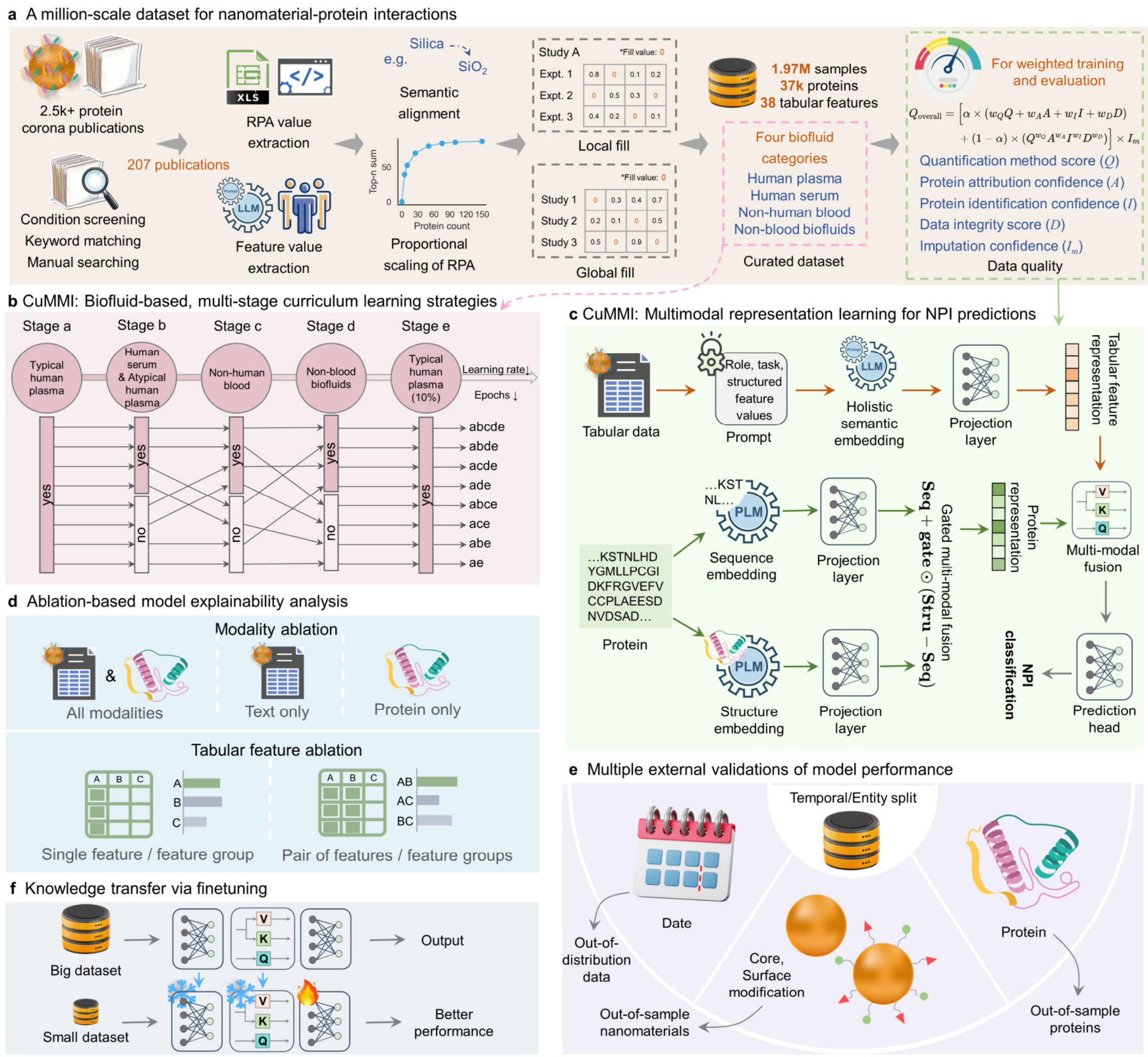

**Figure 1.** Overview and key contributions of this work. (a) Largest-to-date dataset construction for NPI. We retrieved >2,500 articles and retained 207 studies after multi-stage screening, yielding 1.97 million samples across 37,392 proteins. Both local and global fill methods apply specifically to complete experimental treatments at the given resolution—studies reporting only top-n proteins are not imputed. Biofluids are grouped into four source categories for curriculum learning, and a composite quality score (from five quality metrics) is used as sample weights for weighted training and evaluation. (b)

Biofluid-based, multi-stage curriculum learning strategies of CuMMI. We propose a multi-stage curriculum (stages a-e) that progressively increases distributional shift from human plasma to broader and more distant biofluid domains, followed by in-domain refinement. Learning rate and epochs are reduced across stages, and different auxiliary-domain inclusion strategies are compared. (c) Multimodal representation learning framework of CuMMI. Protein sequence and structure representations are extracted from protein pretrained models and combined via gated fusion, while text-encoded experimental context (tabular features) is encoded by a text pretrained model. The fused protein-text embeddings are fed into a prediction head for NPI prediction. (d) Ablation-based model explainability analysis. We conduct modality ablations (protein vs. text-encoded context) and systematic tabular-feature ablations. Both single feature/feature group and pairs of features/ feature groups analyses are performed to quantify contributions. (e) Multiple external validation. Generalization is evaluated under multiple out-of-distribution settings, including splits by timeline, nanomaterial, and protein. (f) Knowledge transfer via finetuning. The projection and multimodal fusion layers are frozen, and only the prediction head is fine-tuned, achieving improved predictive performance on a small dataset.

## Results

### A million-scale NPI dataset lays the foundation for AI modeling

As of April 27, 2025, we performed a comprehensive literature search across Web of Science, PubMed, and Scopus. We extracted structured features from these studies using large language models (LLMs)[29], followed by extensive manual verification. To account for experimental, analytical, and study-specific variability that influences protein corona composition and model generalizability, we curated 37 tabular features spanning six groups: nanomaterial properties, protein source, incubation conditions, separation parameters, proteomic settings, and research purpose. We further categorized biological fluids into four source domains (human plasma, human serum, non-human blood and non-blood biofluids) to capture the major sources of distributional variation in NPI studies. Despite manual harmonization of semantically similar terms, the dataset still contains >100 categories for nanoparticle core and protein-source types and >200 categories for surface modifications, underscoring the diversity and complexity of experimental conditions. For studies reporting only top-ranked proteins, we applied top-n proportional scaling based on reference distributions from complete datasets.

For studies with multiple experimental groups, protein counts were aggregated, and missing proteins were assigned an RPA of zero (called "local fill" method), resulting in 1.63 million samples. Additionally, a "global fill" method was applied by assigning zero to commonly observed proteins not reported in specific studies, adding 0.34 million samples. In addition, we defined five complementary, sample-level data-quality indicators and combined these indicators into a unified quality index (Figure S1), which was used as a training weight during subsequent model construction. Moreover, protein sequences were retrieved from UniProt[30]. The overview of the proposed dataset is shown in Figure 2 and Supplementary Tables S2,S3.

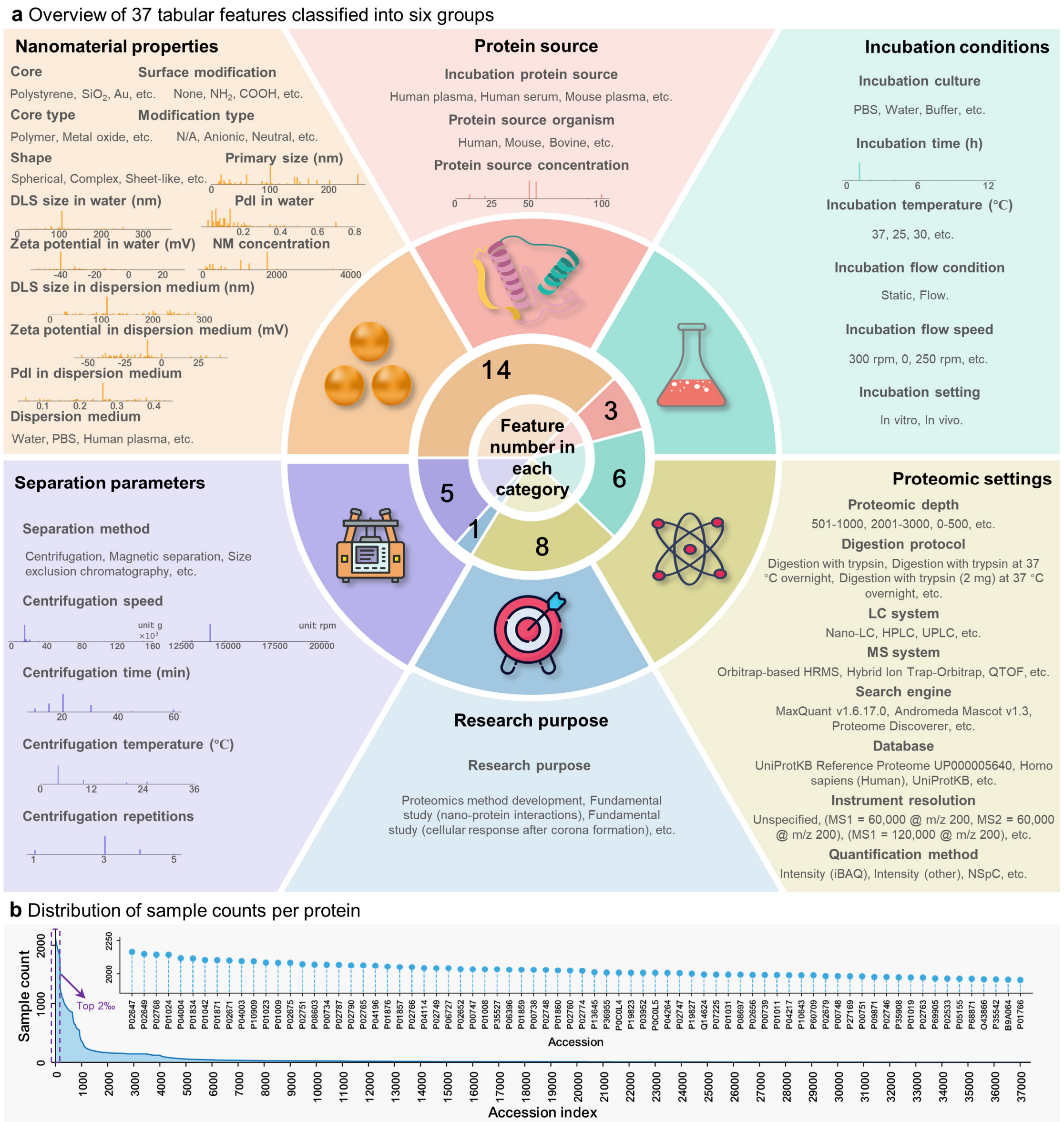

**Figure 2.** Overview of the proposed dataset for nanomaterial-protein interactions. (a) Overview of the 37 tabular features classified into six groups: nanomaterial properties (14 features), protein source (3 features), incubation conditions (6 features), separation parameters (5 features), proteomic settings (8 features), and research purpose (1 feature). For numerical features, the histograms display only the 3rd-97th percentile range to reduce the influence of extreme values. Complete descriptive statistics for all features are provided in the Supplementary Table S2,S3. (b) Distribution of sample counts per protein across 37,392 proteins in the dataset. The overall distribution for all proteins is shown, together

with a zoomed view highlighting the distribution of the top 0.2% most frequently observed proteins.

Multimodal models offer clear advantages for nano-bio interaction prediction by integrating complementary structural, physicochemical, and biological information from both nanomaterials and biosystems[31]. However, multimodal learning poses greater challenges than unimodal approaches due to the need to jointly model heterogeneous modalities, which typically requires larger-scale datasets with well-controlled data quality. In this context, our dataset construction and quality-aware design enable effective use of all available data while reducing the impact of incomplete or low-quality samples, thereby supporting robust multimodal fusion and downstream learning tasks for NPI.

**CuMMI enables effective NPI understanding**

Previous studies have demonstrated that tabular modality data can be used to predict the affinity of specific proteins to various nanomaterial treatments[12], and that the physicochemical properties of protein sequences alone can be employed to predict their affinity to specific nanomaterials[16]. However, the representational capacity of tabular feature is limited. Categorical variables such as nanomaterial type and incubation source are difficult to generalize to unseen categories. Besides, heterogeneous data are challenging to handle; for instance, samples with magnetic separation protocols lack centrifugation parameters, and numerical features may be expressed in diverse units (e.g., concentration in mg/L vs. mol/L, or centrifugation speed in g vs. rpm) even different format (e.g. multiple time intervals with different feature values). Besides, traditional ML models, such as the widely used RF, often perform well on data drawn from the same distribution[32], but show limited generalization to unseen samples. As their learning process relies entirely on the observed training data, they struggle to handle previously unseen inputs or samples with missing values.

To enhance the model's ability to process heterogeneous tabular data and generalize to unseen conditions, we developed CuMMI, a curriculum-guided multimodal representation learning framework. In CuMMI, multimodal representation learning provides inherent generalization capability through pretrained foundation models, while the curriculum learning strategy further strengthens generalization by organizing training from simpler to progressively more challenging scenarios.

Specifically, we leveraged pretrained models for different modalities to learn transferable and generalizable representations. Protein sequence and structure representations were first encoded using the protein language model ESM2 and corresponding structural encoder ESMFold[18], and subsequently integrated through a gated fusion mechanism to form a unified protein representation. Tabular features were converted into structured textual descriptions and represented using Linq-Embed-Mistral[20]. To enable effective cross-modal interaction, the unified protein representation was then fused with the text-based representations via a multimodal fusion architecture based on a multi-head cross-attention mechanism[33], allowing the model to selectively attend to relevant information across modalities (Figure 1c).

Besides, the curriculum learning framework is organized into five sequential stages that progressively increase task difficulty and distributional shift (Figure 1b). Training begins with typical human plasma data to establish core representations, followed by the gradual introduction of atypical plasma and serum data, non-human blood protein corona data, and ultimately non-blood protein corona data, each stage expanding the model's robustness and generalization toward increasingly distant domains. A final fine-tuning stage on a small subset of high-quality human plasma data refines in-domain performance. Collectively, this curriculum comprises a primary source domain and a series of auxiliary intermediate domains, guiding the model from simple to complex learning regimes and supporting robust multimodal generalization.

The performance of CuMMI models on the internal test set is shown in Figure 3. Notably, even on the internal test set, incorporating auxiliary domains in stages b-d yields consistent performance gains (Figure 3a), suggesting that exposure to broader and progressively shifted data distributions can enhance internal accuracy. In contrast, stage e, where the model is fine-tuned on a randomly sampled 10% subset of the target-domain (human plasma) data, leads to a slight drop in internal test performance. This behavior is expected, as fine-tuning on a biased in-domain subset can shift the model toward that subset and away from the original evaluation distribution. Importantly, the auxiliary-domain training and the final stage e refinement are primarily designed to improve performance on unseen data; the specific contributions of each stage are therefore discussed in conjunction with external validation results below.

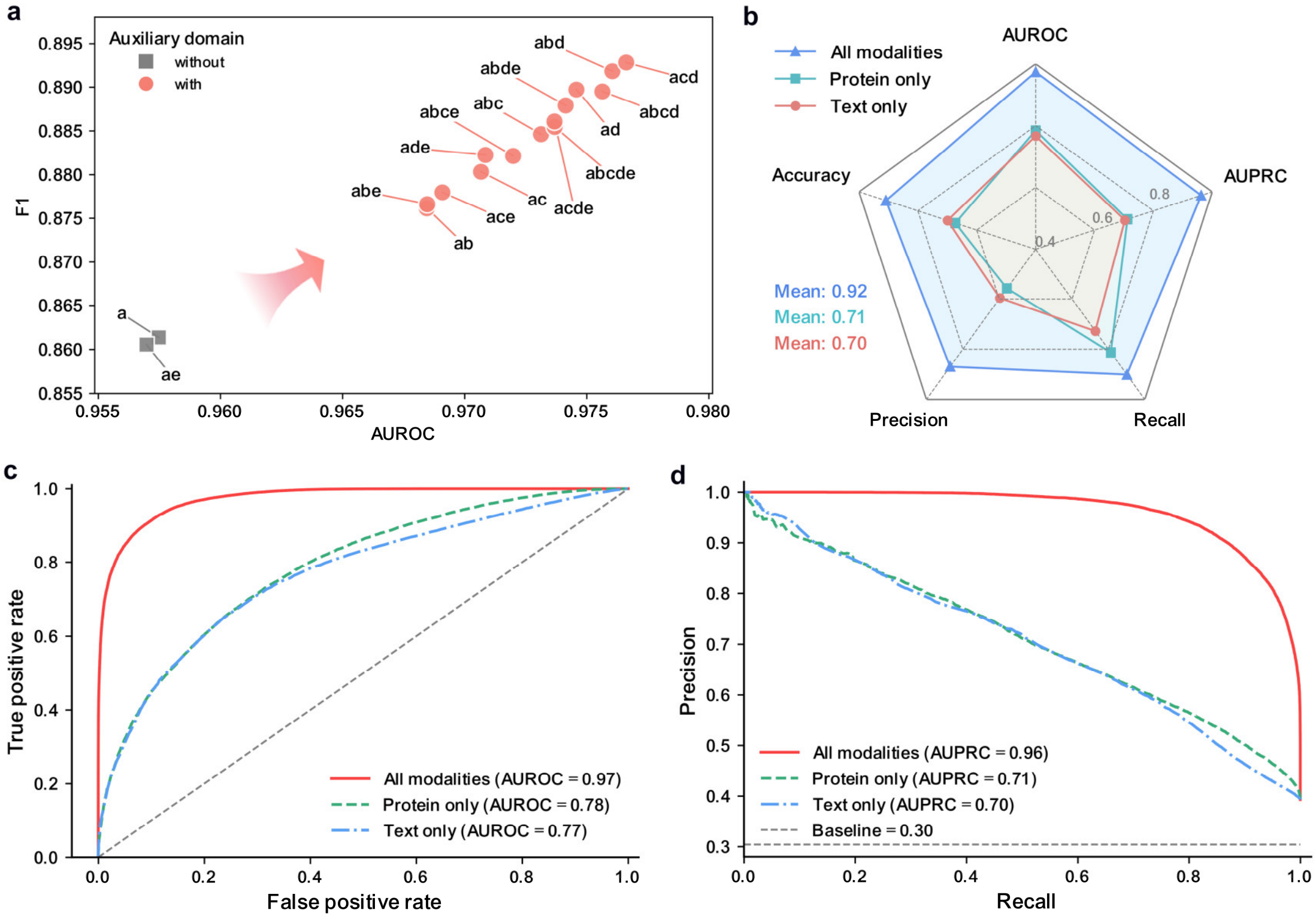

**Figure 3.** Model performance on the internal test set. (a) Performance of different curriculum stage combinations and their intermediate models. (b) Five-stage curriculum-learning performance using all modalities, protein-only modality, and text-only modality. (c) Receiver operating characteristic (ROC) curve and (d) precision-recall (PR) curve for the five-stage curriculum-learning model with all modalities.

Comparisons between multimodal and unimodal models further highlight the benefit of multimodal learning (Figure 3b-d). Across five classification metrics, area under the receiver operating characteristic curve (AUROC), area under the precision-recall curve (AUPRC), recall, precision, and accuracy, the multimodal model consistently outperforms single-modality counterparts, achieving a mean performance of 0.92, compared with 0.71 for the protein-only model and 0.70 for the text-only model (Figure 3b). The protein-only model performs slightly better than the tabular-only model, a trend that is also reflected in the ROC (Figure 3c) and precision-recall curves (Figure 3d), although the difference is modest. This observation underscores the complementary value of integrating nanomaterial context through multimodal fusion, while also highlighting the dominant role of non-specific adsorption processes in NPI.

**Ablation-based tabular feature importance reveals multifactorial effects**

Although protein and text-based pretrained large models offer enhanced generalization capabilities, their vast parameter space, involving billions of parameters, renders model behavior increasingly opaque[34,35]. Nevertheless, providing a certain degree of explainability remains essential. Explainability not only allows for the verification of whether the model has captured meaningful biological or physicochemical patterns, thereby enhancing trustworthiness and transparency, but also enables the extraction of domain-specific insights from the learned representations. Therefore, we employed an ablation-based importance analysis to evaluate the contribution of individual tabular features, and their combined effects and potential interactions (Figure 4).

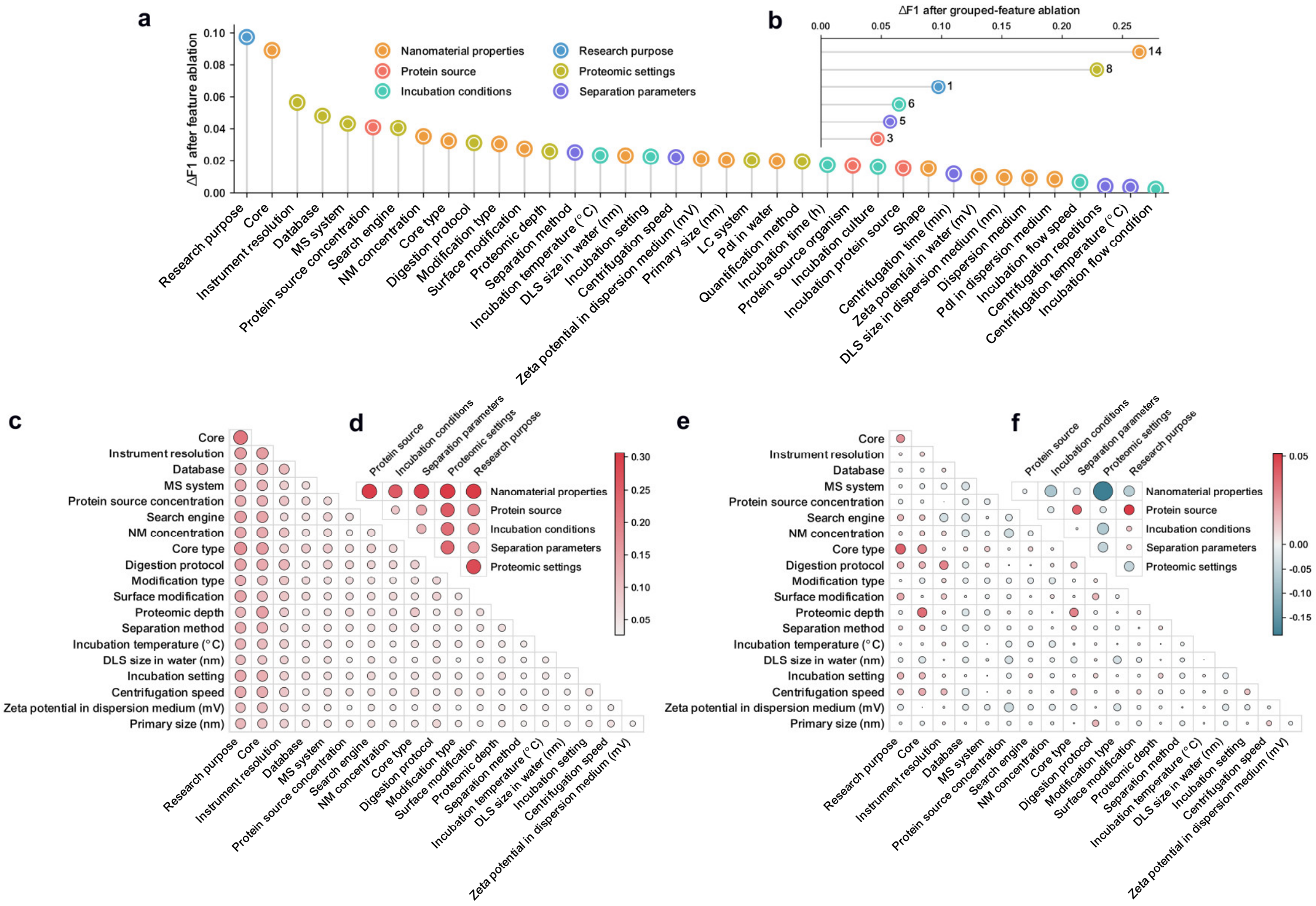

**Figure 4.** Ablation-based importance of tabular features. Effects of (a) single-feature and (b) single feature-group ablations on model predictive performance. Combined effects of (c) two-feature and (d) two feature-group ablations on model predictive performance. Interaction effects of (e) two-feature and (f) two feature-group ablations on model predictive performance, computed by subtracting the summed individual effects (i.e., removing single-feature/single-group contributions).

Ablation analysis shows that model performance is mainly driven by experimental design choices, core and surface modification of nanomaterials, and proteomics-related factors (Figure 4a). At the feature level, research purpose has the strongest influence, indicating that study objectives systematically affect experimental protocols and data reporting, and thereby protein corona composition. Core composition, core type, and surface modification are the most important nanomaterial features, confirming their key role in protein adsorption. It is well-known that surface chemistry of nanomaterials is important factor influencing protein corona formation[36]. Besides, several proteomics-related variables (e.g., instrument resolution, MS system, database, and search engine) also have a strong impact, highlighting the sensitivity of protein corona profiles to analytical workflows. Corona composition characterization is highly dependent on proteomic protocols. In a comparative study, analysis of the same sample by different proteomic facilities yielded only 73 shared proteins out of 4,022 identified (1.8%), highlighting substantial variability across semiquantitative workflows[10]. These procedural details, including protein corona isolation and MS-based characterization, are often regarded as technical steps but have a direct impact on the observed NPI profiles[37], and thus must be carefully considered when interpreting proteomic results. When features are grouped by category, nanomaterial properties and proteomics settings dominate performance contributions, while research purpose ranks next despite being a single feature (Figure 4b). Incubation, separation, and protein-source factors show smaller but still noticeable effects. For instance, previous study indicated incubation conditions and protein source concentration also affected composition of the protein corona[38]. Centrifugation, the most widely used separation method, enables high-throughput processing, but can also alter the composition of the resulting protein corona[39].

To assess combined effects between features, we performed pairwise ablation analysis on the top twenty tabular features to quantify potential synergistic or redundant contributions. Among individual features, the joint ablation of the two most influential variables, nanomaterial core composition and research purpose, resulted in the largest performance degradation (Figure 4c). Notably, the combined effect between core type and research purpose exceeded that of any combinations among the top three features, including instrument resolution with either research purpose or nanomaterial core, underscoring the dominant role of nanomaterial characteristics and study objectives in shaping protein corona profiles. At the feature-group level, pairwise ablation revealed comparable combined effects between nanomaterial properties and separation, proteomic settings, or research purpose (Figure 4d), highlighting the central importance of nanomaterials while also reflecting the systemic and multifactorial nature of NPI. In addition, substantial combined effects were observed between research purpose and proteomic settings, indicating that factors beyond intrinsic nanomaterial and protein properties, such as experimental objectives and analytical workflows, exert non-negligible influences.

After removing individual main effects, the remaining interaction terms reveal distinct synergistic patterns in the combination analysis (Figure 4e). Among individual features, core type and research purpose exhibit the strongest synergistic effect, indicating that their joint contribution exceeds what would be expected from their independent influences alone. Additional synergistic interactions are observed between proteomic depth and nanomaterial core, core type and core composition, and digestion protocol and instrument resolution, reflecting coordinated effects between nanomaterial characteristics and proteomics workflows. At the feature-group level, protein source shows clear synergistic interactions with research purpose and separation parameters, suggesting that biological context and experimental objectives jointly modulate protein corona outcomes (Figure 4f). In contrast, nanomaterial properties and proteomic settings exhibit the highest redundancy, indicating a certain degree of correlation between nanomaterial physicochemical descriptors and proteomics analysis choices. Together, these results highlight that protein corona formation is governed not only by additive effects of individual factors but also by structured interactions between nanomaterials, analytical workflows, and study objectives.

**Three independent external validations demonstrate model generalizability**

We rigorously assessed the generalizability of CuMMI models using three strictly independent external evaluations (Figure 1e), designed to probe performance under realistic and challenging OOD scenarios. These evaluations span temporal, nanomaterial-held-out, and protein-held-out settings, ensuring independence along orthogonal dimensions of the NPI space.

Compared with their effects on the internal test set, auxiliary domains exert a substantially stronger influence on external temporal generalization (Figure 5a). As auxiliary stages are progressively introduced, AUROC on the temporal split increases from approximately 0.58 under conventional training without auxiliary domains (a, ae) to 0.71 for the full multi-stage curriculum (abcde), highlighting improved robustness under prospective deployment.

The weaker linear correspondence between AUROC and F1 in the external setting likely reflects the threshold dependence of F1, as decision thresholds were calibrated on the internal validation set and may be suboptimal under temporal distribution shift. Stage-wise analysis further reveals distinct contributions across evaluation regimes (Figure 5b): while far-domain augmentation (stage d) provides the largest gains on the internal test set, external performance is driven primarily by biologically grounded auxiliary domains, particularly the combination of cross-biofluid (stage b) and cross-species (stage c) augmentation. These results suggest that auxiliary domains capturing realistic biological variability are critical for learning transferable representations that support robust temporal generalization. In the final fine-tuning stage, we evaluated different sampling ratios of typical human plasma data (10-100%) and found that using only 10% achieved the best generalization on the external test set (Supplementary Figure S2). Increasing the proportion did not yield further gains and sometimes degraded performance, likely because excessive fine-tuning on the source domain partially overwrote the generalizable representations learned in earlier curriculum stages.

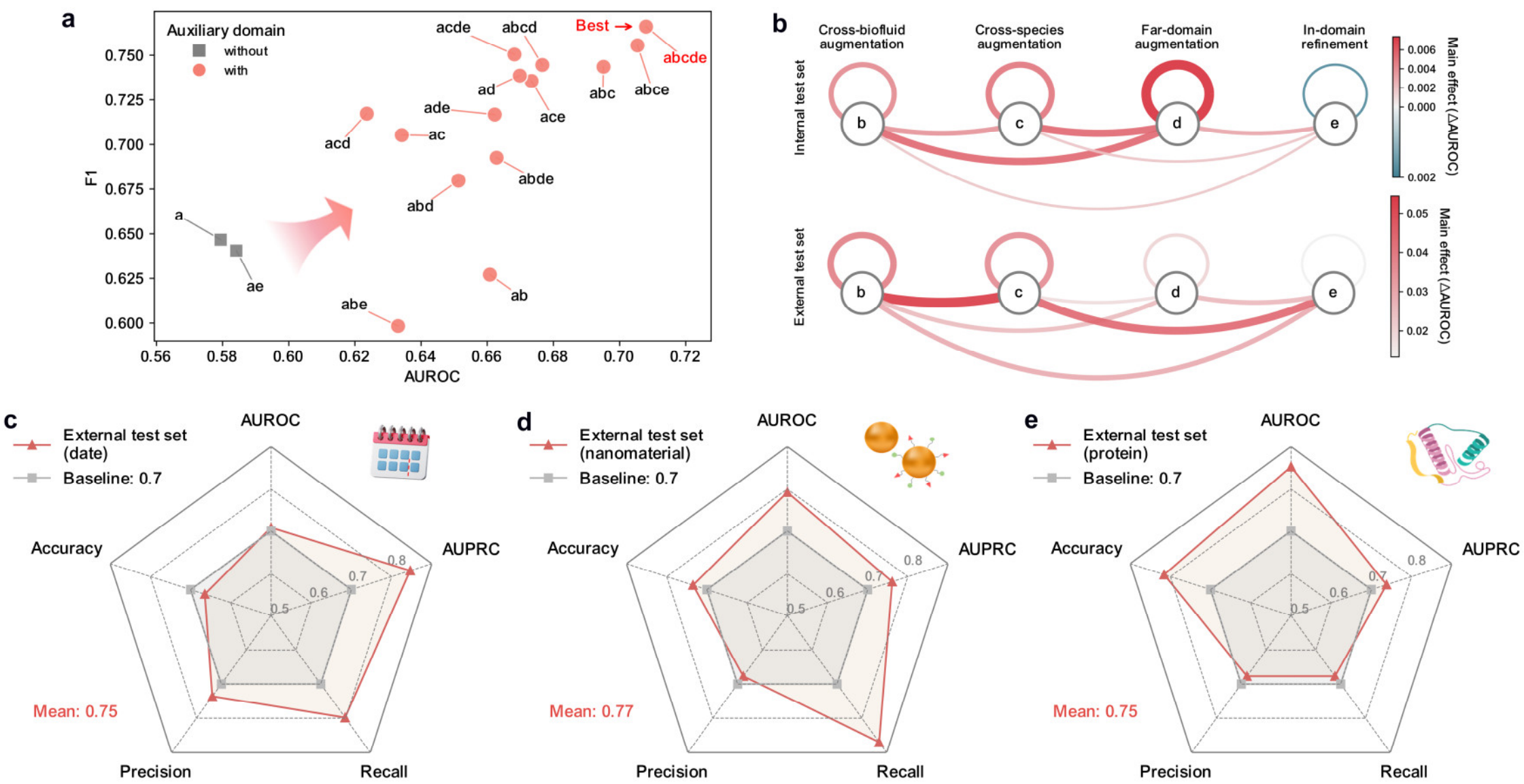

**Figure 5.** Model performance on multiple external test sets and the effects of curriculum-learning stages. (a) Performance of different curriculum-stage combinations and their intermediate models on temporal split test set. (b) Effects of individual stages and two-stage combinations (stages b-e) on model performance on the internal and external test

sets, comparing models with vs. without a given stage (or stage pair). Performance under three independence-preserving evaluations: (c) temporal split, (d) nanomaterial-held-out split, and (e) protein-held-out split.

The results of five classification metrics under the temporal, nanomaterial-held-out, and protein-held-out settings are summarized in Figure 5c-e. Across all three external evaluations, the average performance exceeds 0.75, with threshold-independent metrics, including AUROC and AUPRC, consistently above 0.7, collectively demonstrating robust model generalizability. Building on this capability, the curriculum-guided multimodal representation learning framework enables generalized predictive modeling beyond observed data regimes, achieving the ability to predict interactions for arbitrary combinations of nanomaterials and proteins.

**Knowledge transfer from foundation models enables efficient and accurate modeling**

To evaluate the benefits of knowledge transfer from the established CuMMI model, we compare fine-tuning with training from scratch under data-limited settings (Figure 6). Specifically, all gold nanoparticle samples and a randomly selected subset of 2,359 proteins were fully excluded from the four-biofluid dataset during pretraining, and a curriculum-guided multimodal model was trained on the remaining data. For evaluation, the held-out Au nanoparticle dataset or protein subset was split into 20% test data, with an additional 20% of the remaining data used for validation. From the remaining training pool, varying fractions (10-100%) were randomly sampled to train models either from scratch or via fine-tuning from the pretrained model. Decision thresholds were selected on the validation set and performance was assessed on the test set.

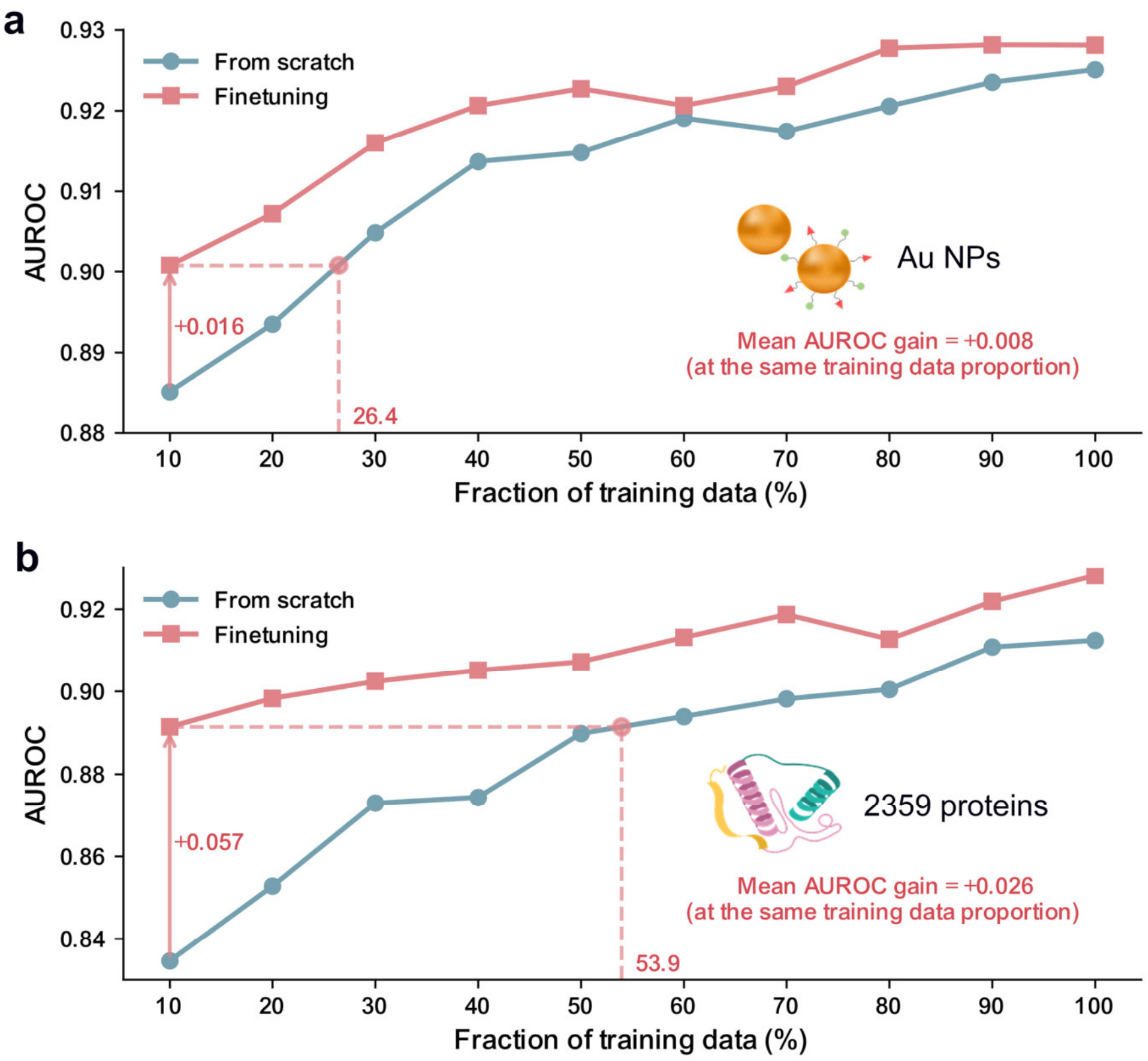

**Figure 6.** Performance comparison between fine-tuning a foundation model and training a model from scratch under different fractions of training data. (a) A held-out gold nanoparticle dataset disjoint from the dataset. (b) A held-out protein subset disjoint from the dataset.

Across all training data fractions, fine-tuned models consistently outperform models trained from scratch. On average, fine-tuning improves performance by 0.008 on the Au nanoparticle dataset and by 0.026 on the held-out protein subset. Notably, the performance gap widens as training data become more limited: at 10% of the available training data, fine-tuning yields improvements of 0.016 for Au nanoparticles and 0.057 for the protein subset. To match the performance achieved by fine-tuning with only 10% of the data, models trained from scratch require approximately 26.4% of the Au nanoparticle data and 53.9% of the protein subset data. Moreover, even when trained on the full dataset, fine-tuned models continue to achieve superior performance. Together, these results demonstrate that curriculum-guided multimodal representation learning enables effective knowledge transfer, substantially improves data efficiency and predictive accuracy, and underscores the practical value of the proposed modeling framework for scalable and generalizable NPI prediction.

## Discussion

Nanomaterials hold great promise in diagnostics and therapies due to their unique advantages in drug delivery[40], imaging[41], and theranostics[42]. Understanding NPI is essential for advancing nanomedicine and ensuring nanomaterial safety in biomedical and environmental contexts[7]. However, the vast combinatorial space of nanomaterials and proteins, combined with sparse data and complex experimental systems, has limited progress in the field. Investigating nanomaterial-protein corona formation solely through experimental approaches remains time-consuming and resource-intensive[43], and conclusions drawn from limited datasets are often difficult to generalize across systems[44]. Existing AI studies often focus on narrow and predefined scenarios, restricting generalizability.

To address these challenges, we propose CuMMI, an integrated framework for nanomaterial–protein interaction modeling that advances the field across three key dimensions: dataset construction, algorithmic design, and learning strategy. We curate, to our knowledge, the largest NPI dataset to date, providing a robust foundation for generalizable prediction. Importantly, each sample is annotated with a quantitative measure of data quality, enabling quality-aware weighted training that fully leverages heterogeneous data while appropriately modulating the influence of lower-confidence observations. At the algorithmic level, we adopt a multimodal representation learning architecture that fuses physicochemical descriptors with representations derived from pretrained text and protein language models, allowing the model to capture the complex, multi-factorial determinants of NPI and to benefit from transferable, high-level representations with strong generalization capacity. At the level of learning strategy, we introduce a biologically informed curriculum learning framework based on biofluids, which organizes training from simpler to more complex interaction scenarios. This progressive exposure increases the diversity of interactions encountered during training and improves robustness to data distribution shifts, thereby enhancing generalization performance.

Reliable prediction of NPI requires explicit consideration of experimental heterogeneity and data quality. Here, we adopt a biologically grounded data organization strategy that defines human plasma as the primary target domain and arranges other biofluids along increasing biological distance, from human serum to non-human blood and non-blood biofluids. We further observe that studies labeled as human plasma are not experimentally uniform, and use the presence of albumin as an empirical indicator of canonical plasma composition; datasets lacking this hallmark are treated as atypical plasma and incorporated into earlier curriculum stages together with human serum data. This stratification provides a principled basis for curriculum-guided training, enabling the model to establish robust representations in a well-defined biological context before progressively adapting to broader and noisier domains. In parallel, we evaluate each interaction sample using five complementary quality metrics capturing quantification rigor, protein attribution and identification confidence, data integrity, and imputation uncertainty, which are aggregated into a composite score used to weight samples during training and evaluation. By attenuating the influence of low-confidence or sparsely reported entries without discarding them, this quality-aware learning strategy improves robustness and supports more trustworthy generalization from heterogeneous literature-derived data.

The strength of CuMMI framework lies in the tight coupling of multimodal representation learning with a biologically informed curriculum. By leveraging pretrained models for protein sequence, structure, and experimental context, the model learns transferable representations that jointly encode intrinsic molecular properties and context-dependent effects. Cross-modal attention enables selective interaction between protein and contextual modalities, allowing relevant information to dominate under varying experimental conditions. Importantly, this multimodal learning is guided by a curriculum that explicitly accounts for distributional heterogeneity in protein corona data. By progressively organizing biofluids from human plasma to increasingly distant domains, the model establishes robust core representations before adapting to more diverse and noisy regimes, followed by in-domain refinement. Together, curriculum-guided domain progression and attention-based multimodal fusion provide a principled mechanism for robust generalization and data-efficient adaptation beyond single-stage or single-modality approaches. Somewhat surprisingly, the inclusion of auxiliary domains still improves performance on the internal test set, albeit marginally (Figure 3a). This observation suggests that generalization and in-domain accuracy are not strictly antagonistic: auxiliary-domain training can simultaneously enhance robustness and accuracy, effectively pushing the boundary of the generalization-accuracy trade-off outward.

As noted earlier, prior studies have predominantly focused on unidirectional relationships, examining either how nanomaterials affect individual proteins or how specific proteins interact with a given nanomaterial. In contrast, our foundation model is trained on millions of nanomaterial-protein pairs that jointly encode protein sequences and experimental context, enabling robust prediction across incomplete inputs, where each sample is missing at least one tabular feature (Supplementary Figure S3), as well as across previously unseen nanomaterials and proteins. These properties make the model particularly well suited for high-throughput screening and data-efficient nanomaterial design. Importantly, the model can be readily fine-tuned using a limited amount of task-specific data to further improve performance, a critical capability given the substantial heterogeneity in experimental protocols and reporting across studies. This combination allows the model to capture generalizable interaction patterns at scale while remaining adaptable to diverse and context-specific experimental settings.

To enhance model explainability and build trust in CuMMI's predictions, we employ the ablation-based measurement method to quantify the relative importance of input modalities and individual features. These evaluations highlight the critical role of integrating both protein-level representations and structured experimental metadata for accurately modeling NPI, an aspect largely overlooked in current research. Additionally, the analysis of tabular feature importance and interactions offers mechanistic insights into the model's decision-making process. The alignment between key features and established domain knowledge supports the model's reliability, while also identifying critical factors in NPI as revealed by our model trained on the largest dataset to date.

Together, our dataset and CuMMI model provide a scalable and transferable computational foundation for systematic exploration of NPI landscapes, with broad implications for nanomedicine design, biosafety assessment, and data-driven nanomaterials discovery. Future work may focus on the underexplored "dark space" of proteomics—proteins that are rarely studied[45] but may play critical roles in biological processes or serve as biomarkers. While our model shows promising generalization to these proteins, further performance improvements will require targeted data augmentation. Additionally, this approach has the potential to significantly accelerate *in vitro* nanomaterial-protein research by reducing reliance on time-consuming and costly experiments, enabling reliable predictions with minimal experiments. For researchers with AI expertise seeking to further improve model performance, promising directions include exploring alternative model architectures and pretraining strategies, incorporating information of protein structure and molecule structure using graph neural network[46], integrating domain knowledge, and enhancing model's intrinsic explainability.

While the present study introduces the most comprehensive dataset to date for NPI and a curriculum-guided multimodal representation learning framework with strong generalization across diverse nanomaterials and proteins, several limitations should be acknowledged. Specifically, the dataset and model focus on the post-interaction phase, a relatively tractable experimental endpoint that does not capture the intrinsically dynamic, non-equilibrium nature of nano-bio interactions. In physiological environments, biomolecular corona formation is a dynamic, multistep process involving initial diffusion to the nanomaterial surface followed by competitive and sequential biomolecule exchange[47,48]. Proteins that first associate with the bare surface may later be displaced as secondary interactions, such as protein-protein associations and surface curvature effects-gain prominence. These dynamic adsorption events are influenced not only by medium composition but also by adsorption sequence and nanoscale geometry, resulting in corona profiles that deviate markedly from classical macroscopic phenomena such as the Vroman effect. Consequently, nanomaterials can selectively enrich rare or unexpected biomolecules from complex biofluids. We suggest that experimental resources currently focused on static observations could be more effectively redirected toward investigating adsorption-desorption kinetics, which is critical for advancing *in vivo* applications of nanomaterials.

Nevertheless, the present work holds significant practical relevance, particularly for post-interaction-based applications such as early and rapid disease diagnostics. Owing to their relative stability and experimental accessibility, hard coronas are widely utilized as a major element of the biological identity of nanomaterials and as a means to target diseases[49]. In clinical settings, where target signals are sparse and embedded in complex biological matrices, the ability to engineer nanomaterials that selectively enrich specific protein biomarkers is particularly valuable. Sparse protein models even outperformed those using basic features and clinical assay data across 52 diseases[50]. Protein adsorption onto nanomaterial surfaces can elicit detectable changes in physicochemical properties, enabling biosensing applications for static detection of disease-relevant signatures [51]. Our dataset and model lay a foundation for zero-shot screening, rational nanomaterial design, and broader deployment in static interaction scenarios, including biosensor development and nano-bio interface engineering, thereby accelerating the translation of nanomaterials into real-world applications.

In summary, CuMMI advances generalizable prediction of NPI through the coordinated development of three key components: a large-scale, systematically curated dataset capturing heterogeneous experimental contexts; a multimodal representation learning architecture that jointly encodes protein sequence, structure, and experiment-related descriptors; and a curriculum-guided learning strategy that explicitly accounts for distributional heterogeneity across biofluids. Together, these elements enable generalizable prediction under incomplete inputs and across unseen nanomaterials and proteins, while supporting intrinsic explainability through modality- and feature-level importance analyses. By reducing reliance on static, high-throughput experimental screening, our framework provides a scalable foundation for *in vitro* nano-bio studies and rational nanomaterial design. Looking forward, extending such models to incorporate stronger structural priors and to capture dynamic interaction processes will be essential for temporally aware modeling and improved in vivo translation.

## Methods

In this section, together with relevant sections in the Supporting Information, we explain the technical details of data collection (Supplementary Section 1), dataset curation (Supplementary Section 2), embedding acquisition (Supplementary Section 3), model establishment and evaluation (Supplementary Section 4), importance assessment (Supplementary Section 5), independent external validations (Supplementary Section 6), and knowledge transfer via fine-tuning (Supplementary Section 7).

### Data collection and dataset curation

As of April 27, 2025, we conducted a comprehensive literature search across Web of Science, PubMed, and Scopus, focusing solely on original research articles. Duplicate entries were removed based on DOI, and articles with insufficient text length or low citation counts were excluded. We applied keyword-based filtering to retain studies containing terms like mass spectrometry, abundance quantification, spectral analysis, and data tables. Given the labor-intensive nature of manual data extraction, we utilized large language models (LLMs) with tailored prompts to streamline the process. Extracted experimental descriptors were manually verified, resulting in 37 curated features categorized into six groups. All numerical values were cross-checked for accuracy against the original publications.

In addition to standard preprocessing steps, such as standardizing capitalization and units, we performed semantic alignment, data cleaning, and imputation. For semantic alignment, we standardized nomenclature across multiple features, such as harmonizing terms like “carbon nanotubes” and “GO” into “carbon,” and categorizing core compositions (e.g., “Au and $Fe_3O_4$” to “Au@$Fe_3O_4$”). Surface modifications were standardized using a domain-specific knowledge vector database and categorized into charge-based types (Anionic, Neutral, Cationic). We mapped shape attributes to six categories: spherical, rod-like, sheet-like, plate-like, polyhedral, and complex. Primary size was standardized in nanometers (nm), with different geometric forms treated accordingly.

Due to inconsistencies in reported RPA values, we estimated RPA using available quantification-related data through normalization or methods like iBAQ. For studies with multiple experimental groups, protein counts were summed across groups, and proteins absent in certain groups were assigned an RPA of zero. For studies reporting only the top-ranked proteins, we applied proportional scaling based on the cumulative abundance from studies with complete protein profiles.

To address data heterogeneity and uncertainty, we defined five sample-level data-quality indicators to assess reliability and completeness of protein abundance records. These included four intrinsic dimensions, quantification method score, protein attribution confidence, protein identification confidence, data integrity score, integrated into a unified overall data quality index. Additionally, imputation confidence was introduced as a modifier to penalize low-specificity imputation values. The final quality index provided a continuous measure of data reliability and was used as a weight during model training and evaluation, ensuring effective utilization of heterogeneous data while reducing the influence of low-confidence entries.

**Embedding acquisition**

We leveraged protein- and text-based pretrained models, which have been trained on massive datasets and have captured a broad range of underlying patterns, enabling them to learn highly generalizable representations across both protein and tabular modalities. For protein-based pretrained models, we inputted the amino acid sequence into ESM2 and ESMFold models to obtain representation vectors. For text-based pretrained models, we designed prompt incorporating task description, background information, and contextual knowledge. By filling sample-specific numerical values into the structured prompts, we obtained representation vectors for experimental context. Inference with large models is time-intensive; therefore, we pre-encoded all unique tabular descriptors and protein sequences to avoid redundant computation for repeated inputs.

**Model establishment and evaluation of CuMMI**

CuMMI consists of two core components: a multimodal representation fusion module and a curriculum-guided learning strategy to enhance generalization across diverse biological contexts. The multimodal fusion module has three blocks: modality-specific projections, gated protein fusion with cross-modal attention, and a prediction head. Structured text, protein sequence, and structure embeddings are projected into a shared 1024-dimensional space. Sequence and structure features are combined using a learned gate, allowing structural information to refine the sequence signal when relevant. The fused protein representation interacts with the experimental-context text through an 8-head cross-attention module (text as query and protein as key/value) and a residual update, with a multilayer perceptron mapping the final feature to a binary prediction. Class imbalance is addressed by up-weighting positive samples, and heterogeneous measurement reliability is accounted for by scaling each sample's loss with a continuous quality weight. The curriculum learning framework consists of five stages that gradually increase task difficulty and distributional shift. Human plasma is split into 70% training, 15% validation, and 15% internal test sets. Auxiliary domains (human serum, atypical human plasma, non-human blood, and non-blood biofluids) have 15% validation splits, with no separate test sets. During curriculum training (Stages b-d), each stage's training set combines auxiliary-domain samples with human plasma to prevent catastrophic forgetting. Model selection at each stage is based on performance across two validation splits: stage-specific and typical human plasma validation splits. For fine-tuning (Stage e), we tested various sampling ratios of typical human plasma data, with 10% yielding the best generalization performance on an independent date-based external test set.

Model performance was evaluated using six classification metrics: AUROC, AUPRC, recall, precision, accuracy, and F1 score. F1 was excluded from radar plots, as it is the harmonic mean of recall and precision. AUROC and AUPRC are threshold-independent, while the remaining metrics depend on the decision threshold determined on the validation set. Sample quality weights were used during evaluation to account for measurement reliability.

**Importance assessment**

We employed an ablation-based approach to assess the contributions of each modality and individual features to model performance. For modality ablation, we trained models using only one modality, either protein embeddings or tabular features, on the same dataset. These models followed an architecture nearly identical to the full model, except that the cross-attention module was replaced with a self-attention mechanism. For the tabular modality, we conducted feature ablation by setting the value of each feature to "Unknown" and re-encoding the input using the same text encoder, and performance degradation on the test set was then measured using the F1 score. To evaluate feature interaction effects, we extended the ablation to feature pairs, where two features were simultaneously masked as "Unknown," and the observed performance loss was compared to the sum of losses from individual feature ablations. Feature-group importance was assessed via group-wise ablation, in which all features within a predefined group were masked as "Unknown" and the resulting F1 drop was measured. Similarly, group-level interactions were evaluated by jointly masking two feature groups and comparing the observed performance drop with the expected additive effect from single-group ablations.

**Three independent external validations**

To rigorously assess model generalizability, we performed three strictly independent external validations under realistic OOD settings: temporal (date-based), nanomaterial-held-out, and protein-held-out splits. For temporal validation, we mimicked prospective deployment by using the five most recent independent human plasma proteomics studies as an external test set and training on all earlier studies to prevent information leakage. For nanomaterial-held-out validation, we defined nanomaterial identity by core-surface modification combinations, filtered qualified combinations to ensure sufficient cross-study support while avoiding overrepresented materials, and then randomly withheld 20% of core-surface pairs by excluding all associated samples from training. For protein-held-out validation, we selected moderately represented proteins (occurring more than five times but fewer than 200 times), randomly withheld 25% of protein accessions, and removed all their interactions from the training set. For each validation setting, models were trained exclusively on the remaining data (withheld samples fully excluded) and then evaluated on the corresponding independent external test set to quantify performance under each OOD scenario.

## Knowledge transfer via fine-tuning

To assess knowledge transfer from foundation models under data-limited conditions, we compared fine-tuning with training from scratch using controlled data splits. All gold nanoparticle samples, constituting approximately 10% of the human plasma dataset and representing a realistic scenario in which a novel nanomaterial is introduced, were fully excluded from the four-biofluid dataset during pretraining. In parallel, following the same protocol as the protein-held-out external validation, a randomly selected subset of 2,359 protein accessions was entirely withheld from pretraining. For downstream evaluation, the held-out Au nanoparticle dataset or protein subset was first split to reserve 20% of the data as an independent test set, and 20% of the remaining samples were further used as a validation set, with varying fractions (10-100%) of the remaining samples used for training. For each data fraction, we trained either a fine-tuned model initialized from the pretrained foundation model or a model trained from scratch using the multimodal represent learning framework. During fine-tuning, modality-specific projection layers and the cross-modality fusion module were frozen and only the prediction head was optimized, whereas models trained from scratch were optimized end-to-end. Decision thresholds were selected on the validation set, and performance was evaluated on the test set to quantify the benefits of knowledge transfer.

## Data availability

The curated dataset is available at https://figshare.com/s/217ff9a5219c280f8c92. Due to the large size of the dataset, it may not be directly opened in Excel, but can be accessed using programming tools such as Python.

## Code availability

Codes for embedding acquisition, model establishment, importance assessment, and finetuning are available at https://github.com/YuHengjie/CuMMI.

**Acknowledgments**

This work is supported by Westlake Education Foundation (grant No. 103110846022301) and China Postdoctoral Science Foundation (grant No. 2024M762941). We thank the reviewers for their valuable comments on data filtering and data quality, which inspired the design of our curriculum learning strategy.

**Author contributions**

H.Yu and Y.J. conceived the study. H.Yu, H.Yang, and S.L. collected and curated the data. H.Yu implemented the code and performed the analyses. K.A.D., Y.Y., Y.J., and H.Yu revised the manuscript. Y.J. and H.Yu acquired funding. All authors discussed the results and approved the final manuscript.